\documentclass[11pt,a4paper]{article}
\usepackage[hyperref]{acl2018}
\usepackage{times}
\usepackage{latexsym}
\usepackage{amsmath}
\usepackage{algorithm}
\usepackage{algpseudocode}
\usepackage{multirow}
\usepackage{url}

\aclfinalcopy 

\title{A Study on Passage Re-ranking in Embedding based Unsupervised Semantic Search}

\author{
Md Faisal Mahbub Chowdhury \\
 IBM Research \\
 {\tt  mchowdh@us.ibm.com} \\\And
 Vijil Chenthamarakshan \\
IBM Research \\
{\tt ecvijil@us.ibm.com} \\\AND
 Rishav Chakravarti \\
IBM Watson \\
{\tt rchakaravarti@us.ibm.com } \\\And  
 Alfio M. Gliozzo \\
IBM Research \\
{\tt gliozzo@us.ibm.com }  
}

\date{}

\begin{document}
\maketitle
\begin{abstract}

State of the art approaches for (embedding based) unsupervised semantic search exploits either compositional similarity (of a query and a passage) or pair-wise word (or term) similarity (from the query and the passage). By design, word based approaches do not incorporate similarity in the larger context (query/passage), while compositional similarity based approaches are usually unable to take advantage of the most important cues in the context. In this paper we propose a new compositional similarity based approach, called variable centroid vector (VCVB), that tries to address both of these limitations. We also presents results using a different type of compositional similarity based approach by exploiting universal sentence embedding. We provide empirical evaluation on two different benchmarks.

\end{abstract}

\section{Introduction}
\label{sec_intro}
Semantic search attempts to improve search accuracy by (indirectly) understanding the searcher's intent and the contextual meaning of words in the query. Projecting words into a non-discrete semantic space for the purposes of computing similarity scores between terms has emerged as a powerful way of augmenting traditional information retrieval (IR) techniques which represent words in a discrete space. 

In existing IR work, compositional similarity based approaches derive (static) centroid vector representations for both the query and the passage from their component words and directly compare these compositional representations. Alternatively, word level similarity approaches accumulate distances between (selected) pairs of words (inside the query and the passage) and then combine these individual scores into an overall similarity score.

Until few years ago, it was common to use centroid vectors in semantic search. In its simplest form, the default/vanilla centroid vector\footnote{Note, in this paper, by ``vector'' we mean dense vector (aka embedding) representation learned instead of the sparse vectors used in term-vector models \cite{salton:1988}.} $\vec{t}$ of a text \emph{t} is the sum of the vectors of its words divided by the number of words in \emph{t}. We call this vector a static centroid vector, since the vector for a passage (in the context of semantic search) does not change in response to the query or the corpus from which the passage is extracted. Similarly, the centroid vector for the query also remains the same regardless of the candidate passage to be compared. \cite{kosmopoulos:2016} proposed construction of centroid by weighting vectors of the words with term frequency (TF) (in the text) and pre-computed inverse document frequency (IDF), to be used for document classification. In other words, like the default centroid vector, all the words in the passage are taken in consideration. Since the TF (for a word in the corresponding passage) and the IDF (pre-computed from a corpus) are constants, the weighted centroid vector is also static (i.e no change in the vector regardless of the query).

The usual way centroid vector is used for semantic search is -- given a query and a text (passage/document) collection, the static centroid vector for the query is compared (usually using cosine similarity) against static centroid of each candidate passage \cite{furnas:1988}. Later, the candidate passages are ranked according to their similarity scores. Note that the centroid for a document/passage remains static regardless of the wording/variation in the search query.

A passage is likely to be much longer than a query.  Hence, the passage can have a number of words which either are not necessarily relevant to the query, or are not required to formulate the actual answer(s) for the query. So, constructing a (static) centroid vector for a passage using all words (in that passage) could lead to suboptimal results because the non-relevant words in the passages will influence the direction of its centroid vector.

The recent growing number of word level similarity based semantic search work are based on a technique called word mover's distance (WMD). It was introduced by \newcite{kusner:2015} for comparing similarity between two texts. It is a measure of the minimum cumulative distance that words from a \texttt{text A} need to be moved to match words from \texttt{a text B}. The lower the distance between the two texts in the vector space, the higher the semantic similarity. The computation of WMD is computationally expensive -- so, a variation called relaxed word movers distance (RWMD) was proposed.

\newcite{brokos:2016} used a combination of weighted centroid vector and RWMD-Q (where the query is \texttt{text A}); while \newcite{kim:2017} used combination of BM25 ranking function \cite{robertson:2009} and WMD to build hybrid models. \newcite{kim:2017} used (maximum cumulative of) cosine similarity for WMD. The aforementioned studies show that RWMD-Q outperforms the (static) centroid vector based approaches.

In this study, we compare the above approaches with a compositional approach that exploits the recently proposed universal sentence embedding models\footnote{It is beyond the scope of this paper to review exiting work on universal sentence embedding models. We refer readers to some of the pioneering work such as \newcite{chemla:2011,kiros:2015,conneau:2017}} and also propose a variable centroid based approach across different benchmark datasets to better understand whether any of them has a significant edge over the other.  The other approach is based on pre-trained universal sentence embedding models. To the best of our knowledge, this work is the 1st attempt to use universal sentence embedding for passage re-ranking.

\section{Proposed Approaches}

\subsection{Variable centroid vector (VCVB)}

\begin{algorithm*}
\caption{Algorithm for computing query focused centroid vector for passages}\label{variable_vector_algo}
\begin{algorithmic}[1]
\State Let $quWords$ := set of all non-stop-words in the given query
\State Let $paWords$ := set of all non-stop-words in the given passage
\State $selectedWords$ := \{\}   
\State $eat$ := predicted expected answer type of the query
\For{Each $qw$ in $quWords$}
		\State $st$ := the most similar word in $paWords$ for $qw$
		\State Add $st$ in the set $selectedWords$
\EndFor

\State $ste$ := the most similar word in $paWords$ for $eat$
\State Add $ste$ in the set $selectedWords$
\State  $\vec{variableCentroidVector}$ := centroid vector for words in $selectedWords$
\State Return  $\vec{variableCentroidVector}$ 
\end{algorithmic}
\end{algorithm*}

We argue that the centroid vector of a passage should not be fixed and should change because of different wording or intention in queries. For example, consider the following queries --

\begin{enumerate}
	\item \textit{Q1:} Who is the President of the United States?
	\item \textit{Q2:} Who is the head of government of the United States?
	\item \textit{Q3:} Where is the President of the United States?
\end{enumerate}

\textit{Q1} and \textit{Q2} have same intention but wording is different. But query \textit{Q3} has both different wording and intention. Now, consider the passage -- 

\begin{quote}
	\textit{P1:} President Trump gave the third longest State of the Union address in Congress in the modern era, surpassed only by President Bill Clinton in 1995 (84 minutes) and in 2000 (88 minutes).
\end{quote}

Regardless of the intent of the above three queries, the weighted centroid vector (as well as the default centroid vector) for the above passage will be the same, i.e. static. 

Unlike vanilla/weighted centroid vector, we propose to use only certain pre-selected words of a text for constructing its centroid vector. This set of pre-selected words for the same text will vary depending on the wording and intention of the question. 

\begin{table*}[ht]
\begin{center}
  \begin{tabular}{|c||c|c|c||c|c|c|}
    \hline
    & \multicolumn{3}{ |c|| }{Insurance QA} & \multicolumn{3}{ |c| }{BioASQ} \\
\hline
    & P@1 & Recall@5 & NDCG@5 & P@1 & Recall@5 & NDCG@5	 \\ \hline \hline
    Solr (BM25) & 0.135 & 0.241 & 0.189 & 0.385 & 0.241 & 0.249 \\ \hline  	 	
    Default centroid & 0.131 & 0.239& 0.183 & 0.278 & 0.181 & 0.183  \\ \hline
    Weighted centroid & 0.128 & 0.227& 0.175 & 0.262 & 0.170 & 0.172   \\ \hline 
    RWMD-Q & \bf 0.184 & \bf 0.283 & \bf 0.233 & \bf 0.404 & 0.230 & 0.246 \\ \hline
    \textbf{Universal sentence embedding} & 0.128 & 0.233 & 0.179  & 0.262 & 0.167 & 0.164\\ \hline
    \textbf{Variable centroid} & 0.183 & 0.281 & 0.231  & 0.402 & \bf 0.232  & \bf 0.248\\ \hline
  \end{tabular}
\end{center}
\caption{\label{w2v-table} Results of different compositional and word based embedding approaches.}
\end{table*}

For \textit{Q1}, \textit{Q2} and \textit{Q3} mentioned in Section  \ref{sec_intro}, the $selectedWords$ from $P1$ using the above algorithm would be something like the following --

\begin{itemize}
	\item  $selectedWords$ for \textit{Q1}: \textit{\{trump, president, union, state\}}
	\item  $selectedWords$ for \textit{Q3}: \textit{\{trump, president, congress, union, state\}}
		\item  $selectedWords$ for \textit{Q3}: \textit{\{congress, president, union, state\}}
\end{itemize}

As we can see, the sets of $selectedWords$ for the same passage are different for different queries and, hence, the resulting centroid vectors will also vary with respect to the query.

Lets consider another passage --
\begin{quote}
	\textit{P2:} President Trump will give the State of the Union address on January 31.
\end{quote}

For \textit{Q1}, the $selectedWords$ from $P2$ using the above algorithm would be \textit{\{trump, president, union, state\}}. Note, even if wording for passage $P1$ and $P2$ are different, we will have same (variable) centroid vectors from them for a particular question. So, the above algorithm will lead us to conclude that these two passages should have the same ranking for this particular question ($Q1$), which is correct.

By design, the proposed variable centroid vector maximizes the likelihood of similarity between a given passage and the query. In other words, our goal is to judge the semantic importance of a passage from the perspective of the query and not based on what the passage is in general about.

Overall, the proposed approach works as follows. Given a query \texttt{Q}, a search engine returns a set of ranked passages (from indexed documents) using an exiting state-of-the-art retrieval function. These are the first pass search results. Then, the variable centroid vector for each of the top \texttt{N} ranked passages (with respect to \texttt{Q}) is computed. These top \texttt{N} passages are re-ranked according to the similarity between the variable centroid vectors and the centroid vector of \texttt{Q}. If two passages have identical similarity scores, the scores of the passages from the first pass is used to break the tie.

We believe that the step \texttt{6} (and subsequently, steps \texttt{9} and \texttt{10}) in the algorithm would be effective for factoid type questions. Because, by design, these steps would make the proposed variable centroid vector of a likely passage (that contains the actual answer or entities having similar semantic type of the answer) rank higher than an unlikely passage (i.e. a passage that has neither the answer nor any entity of same type).

\subsection{Universal sentence embedding based approach}

The Universal Sentence Encoder~\cite{DBLP:journals/corr/abs-1803-11175} was developed to encode sentences into embedding vectors with the objective of improving transfer learning performance. These vectors are hence designed to be useful for a variety of downstream NLP tasks. For our experiments, we use pretrained embeddings based on a Deep Averaging Network (DAN)\cite{Iyyer:Manjunatha:Boyd-Graber:III}, available for download from TF Hub\footnote{https://tfhub.dev/google/
universal-sentence-encoder/1}. We compute the sentence embedding for the question phrase and compute its cosine similarity with the sentence embeddings of all the candidate answer phrases. This similarity metric is used to re-rank the answer phrases. 

\section{Datasets}

We use two publicly available question answering datasets, each from a different domain: Insurance QA\footnote{\url{https://github.com/shuzi/insuranceQA}} \cite{feng:2015} and  BioASQ \cite{tsatsaronis2015overview}. These datasets come with natural language questions along with a set of relevant answer ids from the corpus, where the answers correspond to passages.

The Insurance QA corpus consists of 27,413 answers written as part of previously submitted support queries. Since our approach is unsupervised, we did not make use of the training question set. The test set contains 2,000 questions. 

In the case of BioASQ, the corpus consists of titles and abstracts from 14,939,692 MEDLINE/PubMED citation records. We use the 1,307 labelled questions made available for evaluation of Task 4b.

In both cases, as first pass results, we identify top 20 candidate passages per question using a Dirichlet smoothed language model similarity score \cite{Zhai:2001:SSM:383952.384019} (implemented in Solr\footnote{Using LM Dirichlet similarity.}) applied over the entire corpus. The value \textit{20} is chosen arbitrarily.

\section{Experimental Results}

In all the experiments, we used cosine similarity for selecting words from passages. The Insurance QA dataset does not contain any factoid questions; while BioASQ has a limited number of factoid questions (checked manually with random selection). So, we skip steps 4, 9 and 10 in Algorithm \ref{variable_vector_algo} in our experiments reported below.

Pre-trained Google word embeddings\footnote{https://code.google.com/archive/p/word2vec/} \cite{mikolov:2013}, i.e. a neural network based distributional semantics approach, were used for all the centroid based approaches as well as RWMD-Q to derive word vectors.  

Usage of the proposed variable centroid vector improves the results significantly with respect to that of the first pass results in both datasets. The proposed approach outperforms both the default and weighted centroid vector based approaches as well as the scores derived using the universal sentence embeddings. 
However, neither RWMD-Q nor variable centroid performs clearly better than the other. Variable centroid gives the best results for Recall@5 and NDCG@5 on the BioASQ dataset, while RWMD-Q works best on Insurance QA dataset. However the performance of both these methods are comparable. 

\section{Conclusion and Future Work}

In this paper, we proposed an approach to compute centroid vector for passages that are not fixed, but varies according to the wording and intention of the given query. We showed experimentally the variable centroid vectors computed using our approach outperforms exiting static centroid vector based approaches. In other words, we proposed a state-of-the-art compositional similarity based approach.

The difference of results between VCVB, the proposed compositional similarity based approach, and RWMD-Q, the state-of-the-art word similarity based approach, is statistically insignificant. 
  
The results also suggest that universal sentence embedding models do not provide an improvement over any of the other approaches.

As a future work, we would like to test on datasets that contain a significant number of factoid questions in order to evaluate the utility of adding answer type detection in these semantic re-ranking strategies. We believe that proposed VCVB would have a better impact on such datasets. 

\bibliography{se2018}
\bibliographystyle{acl_natbib}

\end{document}